%% file: samplepaper.tex
\documentclass[runningheads]{llncs}
\usepackage[T1]{fontenc}
\usepackage{amsmath}       
\usepackage{amssymb}       
\usepackage{graphicx}      
\usepackage{booktabs}      
\usepackage{multirow}      
\usepackage{cite}          
\usepackage{hyperref}      
\usepackage{algorithm}     
\usepackage{algorithmic}   
\usepackage{xcolor}        
\usepackage{orcidlink}
\usepackage{xspace}
\usepackage{tikz}
\usepackage{fontawesome5}
\usepackage{microtype}

\usetikzlibrary{shapes,arrows,positioning,calc}
\newcommand{\methname}{ClipSum\xspace}
\begin{document}
%
\title{Multimodal Abstractive Summarization of Instructional Videos with Vision-Language Models}
\titlerunning{ClipSum: Multimodal Summarization of Instructional Videos}
%
\author{Maham Nazir\inst{1}\orcidlink{0009-0004-1832-297X} \and
Muhammad Aqeel\inst{2}\orcidlink{0009-0000-5095-605X}\textsuperscript{\dag} \and
Richong Zhang\inst{1}\orcidlink{1111-2222-3333-4444}\textsuperscript{*} \and
Francesco Setti\inst{2}\orcidlink{0000-0002-0015-5534}}
\authorrunning{Maham Nazir et al.}
%
\institute{Beihang University, Beijing, China \and
University of Verona, Italy \\
\email{muhammad.aqeel@univr.it} \\
{\small \textsuperscript{*}Corresponding author \quad \textsuperscript{\dag}Contact author}
}
\maketitle              

\input{sec/0_abstract}
\input{sec/1_intro}
\input{sec/2_relatedwork}
\input{sec/3_method}
\input{sec/4_result}
\input{sec/5_ablation}
\input{sec/6_conclusion}
\newpage
\bibliographystyle{splncs04}
\bibliography{mybibliography}
%




\end{document}

%% file: sec/0_abstract.tex
\begin{abstract}
Multimodal video summarization requires visual features that align semantically with language generation. Traditional approaches rely on CNN features trained for object classification, which represent visual concepts as discrete categories not aligned with natural language. We propose \methname{}, a framework that leverages frozen CLIP vision-language features with explicit temporal modeling and dimension-adaptive fusion for instructional video summarization. CLIP's contrastive pre-training on 400M image-text pairs yields visual features semantically aligned with the linguistic concepts that text decoders generate, bridging the vision-language gap at the representation level. On YouCook2, \methname{} achieves 33.0\% ROUGE-1 versus 30.5\% for ResNet-152 with 4$\times$ lower dimensionality (512 vs. 2048), demonstrating that semantic alignment matters more than feature capacity. Frozen CLIP (33.0\%) surpasses fine-tuned CLIP (32.3\%), showing that preserving pre-trained alignment is more valuable than task-specific adaptation. \href{https://github.com/aqeeelmirza/clipsum}{\textbf{GitHub Page}}
\keywords{Multimodal Abstractive Summarization \and Vision-Language Models \and CLIP \and Instructional Video Understanding \and Cross-Modal Fusion}
\end{abstract}

%% file: sec/1_intro.tex
\section{Introduction}
\label{sec:introduction}

\begin{figure}[t]
    \centering
    \includegraphics[width=\textwidth]{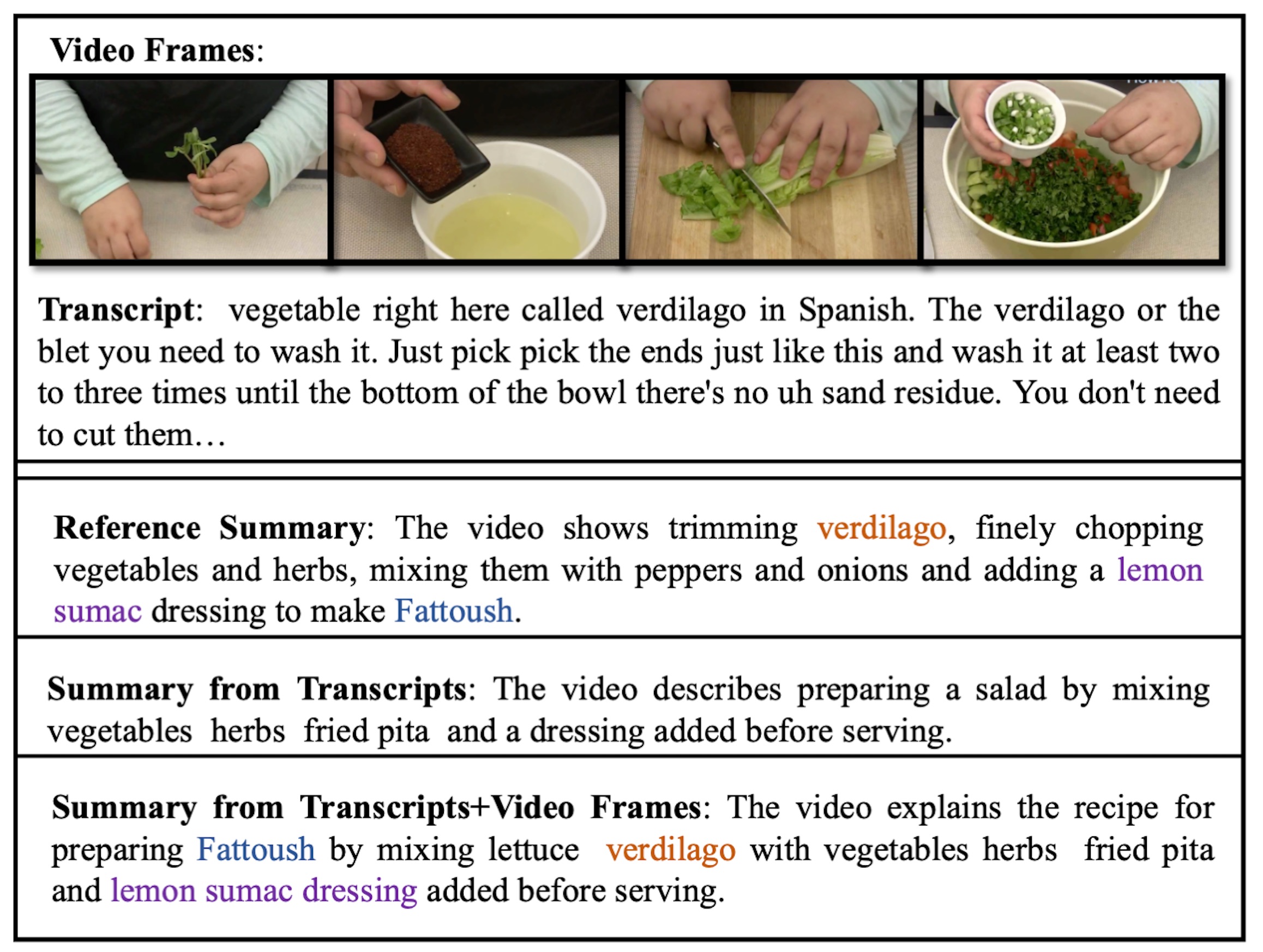}
    \caption{An example of multimodal abstractive summarization. Input is the video frames from the cooking tutorials and their captions. The {\ldots} represents unimportant omitted text. Some emphasized elements (e.g., \textbf{verdilago} or \textbf{Fattoush salad preparation steps}) exist only in the visual signal. The summaries with and without visual data are illustrated in comparison with the human-generated reference summaries.}
    \label{fig:Introduction-figure}
\end{figure}

Instructional videos have become a primary medium for knowledge sharing across platforms like YouTube. However, their length makes it challenging for users to quickly assess content relevance without watching entire videos. Automatic video summarization addresses this need by generating concise textual summaries that capture essential information. While text-only summarization methods have proven effective for articles and transcripts, ``videos contain crucial visual information, ingredient appearances, cooking techniques, and plating presentations'' that textual descriptions alone cannot fully convey. This raises a fundamental question:~\emph{how can we effectively leverage both visual and textual modalities to generate high-quality abstractive summaries from instructional videos?} Figure~\ref{fig:Introduction-figure} illustrates this challenge: visual information such as specific ingredients (e.g., verdilago) and preparation techniques for Fattoush salad are lost when relying solely on textual captions, resulting in less informative summaries compared to multimodal approaches.

Effective multimodal summarization requires visual features that align semantically with language generation. Traditional approaches rely on CNN features trained for object classification, which represent visual concepts as discrete categories (e.g., ``knife,'' ``tomato,'' ``cutting board'') rather than semantic concepts aligned with natural language. This categorical representation creates a semantic gap when interfacing with language generation models, which operate in a continuous semantic space where concepts like ``sautéing'' or ``caramelizing'' are represented through distributional semantics. For instance, when summarizing a cooking procedure, CNN features may independently detect ``chicken,'' ``sauce,'' and ``pan'' as separate object categories but fail to capture the unified procedural concept of ``coating chicken with sauce,'' an action naturally understood through its linguistic description.

Vision-language pre-trained models like CLIP~\cite{radford2021learning} offer a solution to this semantic gap. CLIP learns visual representations through contrastive training on 400 million image-text pairs, aligning visual concepts directly with their natural language descriptions. This vision-language alignment creates representations inherently suited for text generation, where understanding ``slicing tomatoes'' as a coherent visual-semantic concept is more valuable than detecting ``knife'' and ``tomato'' as separate objects.

To leverage this alignment, we propose \textbf{\methname{}}, a multimodal framework for instructional video summarization that integrates frozen CLIP visual features with BART through explicit temporal modeling and dimension-adaptive cross-modal fusion. We validate \methname{} on YouCook2~\cite{zhou2018towards}, a large-scale instructional cooking video dataset where understanding procedural actions and ingredient interactions is essential for generating useful summaries.

Our main contributions are summarized as follows:
\begin{itemize}
    \item We propose \methname{} three architectural components: a frozen vision-language encoder that preserves CLIP's pre-trained alignment while learning only lightweight adaptation layers; explicit temporal modeling through positional encodings and Transformer processing to capture procedural action sequences; and dimension-adaptive cross-modal fusion that integrates CLIP's compact 512-dimensional semantic features with BART's 768-dimensional text representations.
    
    \item We establish design principles for leveraging vision-language models in generation tasks: freeze encoders to preserve pre-trained alignment, model temporal dynamics explicitly for procedural understanding, and fuse modalities at mid-layers for optimal integration.
    
    \item We discover that freezing CLIP outperforms fine-tuning, revealing that preserving large-scale pre-trained alignment is more valuable than task-specific adaptation, a counterintuitive finding with broad implications for vision-language model deployment.
\end{itemize}

Beyond architecture and findings, it \methname{} offers practical advantages: freezing the visual encoder not only preserves CLIP's pre-trained alignment but also reduces computational cost and memory requirements, making multimodal summarization more accessible for resource-constrained settings. Our work demonstrates that vision-language models, when properly adapted through frozen encoders and explicit temporal modeling, provide effective representations for multimodal text generation tasks.

%% file: sec/2_relatedwork.tex
\section{Related Work}
\label{sec:relatedwork}

\subsection{Video Summarization and Pre-trained Language Models}
Video summarization has evolved from extractive methods that select keyframes based on visual saliency~\cite{song2015tvsum,gygli2014creating,zhang2016video} to abstractive approaches that generate natural language descriptions~\cite{palaskar2019multimodal,zhu2020multimodal, song2020video}. Pre-trained sequence-to-sequence models such as BART~\cite{lewis2020bart}, which combines bidirectional encoding with autoregressive decoding through denoising pre-training, have become powerful foundations for text generation tasks and including multimodal summarization.
For multimodal summarization, effective fusion of visual and textual information is essential. Early approaches concatenated visual and text features~\cite{li2017multi,hori2017attention}, while more sophisticated methods employed cross-modal attention mechanisms~\cite{libovicky2017attention,palaskar2019multimodal}. Recent work has explored injecting visual features at different layers of pre-trained language models, with mid-layer fusion strategies~\cite{yu2021vision} demonstrating an effective balance between modality-specific processing and cross-modal integration. These approaches typically employ CNN features from models like ResNet~\cite{he2016deep} pre-trained on ImageNet classification, which constrains visual representations to a closed set of object categories rather than the open-vocabulary semantics required for free-form summary generation.

\subsection{Vision-Language Pre-training}
The emergence of large-scale vision-language pre-training has fundamentally changed visual representation for multimodal tasks. Traditional CNN features from ImageNet~\cite{deng2009imagenet} are optimized for distinguishing object categories but lack explicit alignment with natural language. In contrast, vision-language models learn joint embeddings that bridge visual and textual modalities.
CLIP~\cite{radford2021learning} pioneered this approach through contrastive learning on 400 million image-text pairs from the internet. Unlike ImageNet's 1.2 million labeled images with 1,000 categories, CLIP learns from natural language supervision at web scale, enabling the understanding of diverse visual concepts described in free-form text. The model uses contrastive learning to align image and text encoders, resulting in visual features semantically aligned with text descriptions. This alignment makes CLIP features particularly suitable for tasks requiring vision-language interaction. Subsequent works like BLIP~\cite{li2022blip} and ALBEF~\cite{li2021align} have further improved vision-language alignment through unified architectures and momentum distillation, though CLIP remains widely adopted for its effectiveness and simplicity.
Building on these advances in vision-language pre-training and multimodal fusion strategies, we propose \methname{}, a framework that leverages frozen CLIP features with explicit temporal modeling for instructional video summarization. Our approach preserves large-scale vision-language alignment while adapting to the temporal dynamics of procedural videos through lightweight architectural components.

%% file: sec/3_method.tex
\section{\methname{} Pipeline}
\label{sec:method}

\begin{figure}[t!]
    \centering
    \includegraphics[width=\textwidth]{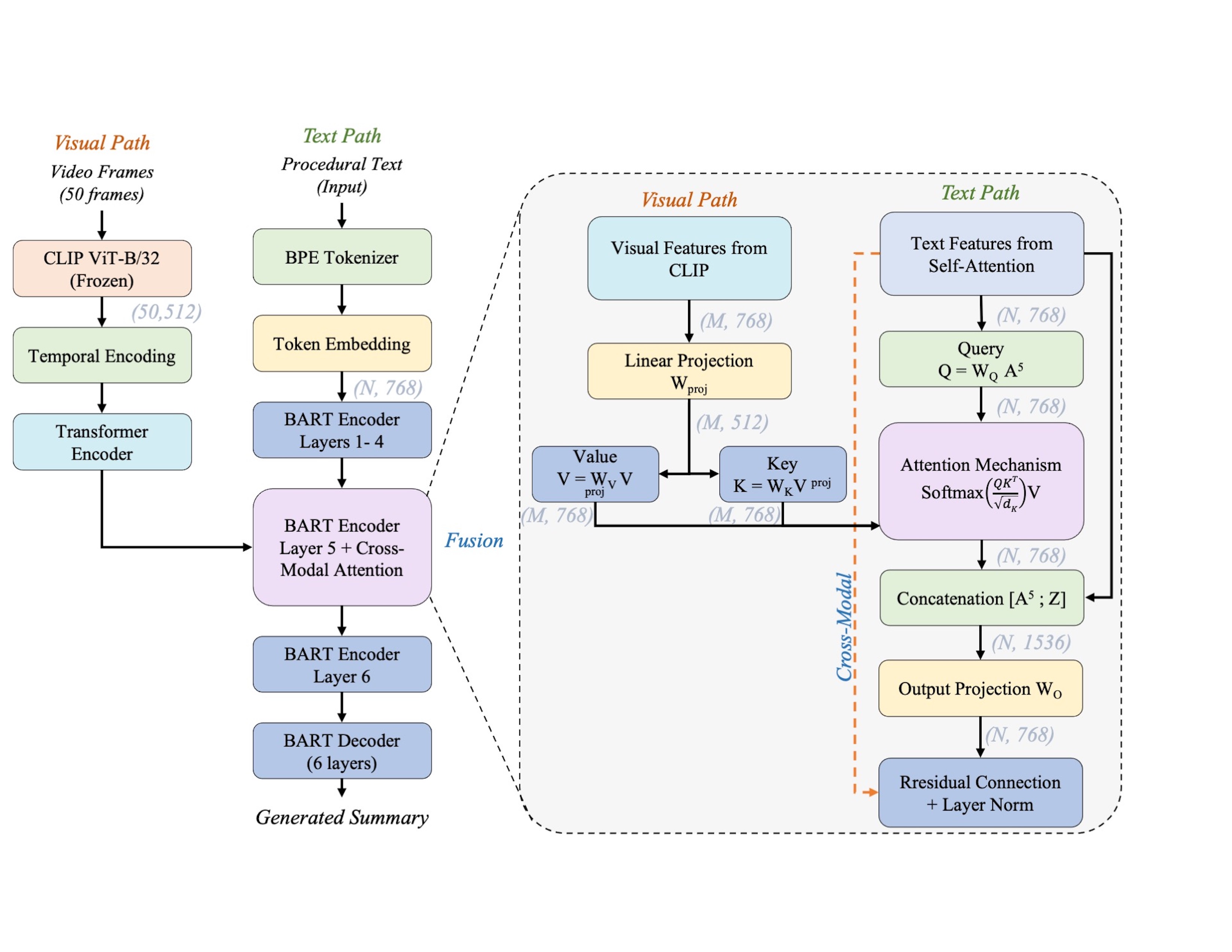}
    \caption{Architecture overview of \methname{}. Video frames are encoded through frozen CLIP ViT-B/32 to obtain 512-dimensional visual features, while procedural text is processed through BART encoder layers 1-4 (768-dim). Visual features are linearly projected and fused with text representations via cross-modal attention at encoder layer 5 (details shown in inset). The cross-modal attention mechanism computes queries from text features and keys/values from projected visual features, producing attended representations that are concatenated, projected, and combined via residual connection. The fused multimodal representations are refined in encoder layer 6 before the BART decoder autoregressively generates the abstractive summary.}
    \label{fig:architecture-figure}
\end{figure}

We propose \methname{}, a multimodal abstractive summarization framework that leverages frozen vision-language features with explicit temporal modeling for instructional videos. Our framework employs CLIP~\cite{radford2021learning} visual encoders pre-trained with natural language supervision, providing semantically meaningful representations aligned with text descriptions. These visual features are integrated with BART~\cite{lewis2020bart} text encoder-decoder through dimension-adaptive cross-modal attention, enabling the model to generate vision-aware summaries while maintaining strong language generation capabilities. Figure~\ref{fig:architecture-figure} illustrates the overall architecture of \methname{}.

\subsection{Problem Formulation}

Given an instructional video $\mathcal{V}$ and its corresponding procedural step descriptions $\mathcal{T} = \{t_1, t_2, ..., t_K\}$, our goal is to generate a concise abstractive summary $\mathcal{S}$ that captures the essential procedure demonstrated in the video. The procedural steps $\mathcal{T}$ provide textual descriptions of actions, while video frames offer complementary visual information about objects, scenes, and spatial relationships. The key challenge lies in effectively fusing these modalities to produce summaries that are both linguistically fluent and grounded in visual content.

\subsection{Frozen Vision-Language Feature Extraction}

\textbf{Visual Encoding with Frozen CLIP.}
For each video $\mathcal{V}$,
we uniformly sample $M=50$ frames to capture the temporal progression of the instructional procedure. Each sampled frame $f_i$ is resized to $224 \times 224$ pixels and encoded through CLIP's visual encoder, which is a Vision Transformer (ViT-B/32)~\cite{dosovitskiy2021image} that divides images into $32 \times 32$ pixel patches, pre-trained on 400 million image-text pairs through contrastive learning.

The CLIP visual encoder partitions each frame into patches of size $32 \times 32$ (yielding 49 patches per image) and processes them through 12 transformer layers with multi-head self-attention. Critically, we freeze all CLIP parameters during training to preserve its large-scale vision-language alignment. 
The encoder outputs a 512-dimensional feature vector from the final [CLS] token:
\begin{equation}
\mathbf{v}_i = \text{CLIP}_{\text{visual}}(f_i) \in \mathbb{R}^{512}
\end{equation}

This produces a sequence of visual features $\mathbf{V} = [\mathbf{v}_1, \mathbf{v}_2, ..., \mathbf{v}_M] \in \mathbb{R}^{M \times 512}$ for the video. These features encode semantic relationships aligned with text, capturing holistic concepts (e.g., ``slicing tomatoes'') rather than isolated object detections.

\textbf{Explicit Temporal Modeling.}
To preserve temporal ordering and capture procedural dynamics, we add learnable positional encodings to the visual features:
\begin{equation}
\tilde{\mathbf{V}} = \mathbf{V} + \mathbf{PE}_{\text{visual}} \in \mathbb{R}^{M \times 512}
\end{equation}
where $\mathbf{PE}_{\text{visual}} \in \mathbb{R}^{M \times 512}$ are learned positional embeddings. We then process these temporally encoded features through a 2-layer Transformer encoder with 4 attention heads and a feed-forward dimension of 1024 to model inter-frame dependencies. This explicit temporal modeling is essential for understanding procedural action sequences such as ``marinate, then fry, then coat'' in cooking videos, where the order of operations determines the final outcome.

\subsection{Text Encoder-Decoder with BART}
We adopt BART-base~\cite{lewis2020bart} as our text encoder-decoder backbone. BART is pre-trained with a denoising objective that learns to reconstruct corrupted text, making it particularly effective for abstractive generation. The architecture consists of a 6-layer bidirectional encoder and a 6-layer autoregressive decoder, with each layer having a hidden dimension $d=768$.

\textbf{Text Encoding.}
The input procedural steps $\mathcal{T}$ are tokenized using BART's byte-pair encoding (BPE) tokenizer, producing a token sequence of length $N$, $\mathbf{X} = [x_1, x_2, \ldots, x_N]$. Each token is embedded and combined with positional encodings:
\begin{equation}
\mathbf{H}^0 = \text{Embed}(\mathbf{X}) + \mathbf{PE}_{\text{text}} \in \mathbb{R}^{N \times 768}
\end{equation}

The encoder processes these embeddings through 6 transformer layers, where each layer $\ell$ contains multi-head self-attention (MSA) and feed-forward network (FFN) sub-layers with residual connections and layer normalization (LN):
\begin{align}
\mathbf{A}^\ell &= \text{LN}(\text{MSA}(\mathbf{H}^{\ell-1}) + \mathbf{H}^{\ell-1}) \\
\mathbf{H}^\ell &= \text{LN}(\text{FFN}(\mathbf{A}^\ell) + \mathbf{A}^\ell)
\end{align}

\textbf{Text Decoding.}
The decoder generates the summary $\mathcal{S}$ autoregressively, using masked self-attention to attend to previously generated tokens and encoder-decoder attention to incorporate encoded information. During training, we use ``teacher forcing,'' where ground-truth tokens are provided as input. The final layer outputs a probability distribution over the vocabulary for each position.

\subsection{Dimension-Adaptive Cross-Modal Fusion}
To effectively integrate visual information while preserving BART's pre-trained language generation capability, we inject visual features through cross-modal attention at encoder layer 5. This mid-layer position allows the model to first process modality-specific information in early layers before fusing at an intermediate stage, then refining the multimodal representation in the final encoder layer.

\textbf{Visual Feature Projection.}
Since CLIP features (512-dim) differ in dimensionality from BART hidden states (768-dim), we project the visual features into the text feature space:
\begin{equation}
\mathbf{V}^{\text{proj}} = \mathbf{W}_{\text{proj}} \tilde{\mathbf{V}} \in \mathbb{R}^{M \times 768}
\end{equation}
where $\mathbf{W}_{\text{proj}} \in \mathbb{R}^{768 \times 512}$ is a learnable projection matrix. This dimension-adaptive design accommodates CLIP's compact semantic representations.

\textbf{Cross-Modal Attention Mechanism.}
At encoder layer 5, after the self-attention sub-layer, we insert a cross-modal attention mechanism that allows text representations to query relevant visual information. Given text representations $\mathbf{A}^5 \in \mathbb{R}^{N \times 768}$ from the self-attention sub-layer and projected visual features $\mathbf{V}^{\text{proj}}$, we compute queries, keys, and values:
\begin{align}
\mathbf{Q} &= \mathbf{W}_Q \mathbf{A}^5 \in \mathbb{R}^{N \times 768} \\
\mathbf{K} &= \mathbf{W}_K \mathbf{V}^{\text{proj}} \in \mathbb{R}^{M \times 768} \\
\mathbf{V}_{\text{attn}} &= \mathbf{W}_V \mathbf{V}^{\text{proj}} \in \mathbb{R}^{M \times 768}
\end{align}

The attention mechanism computes similarity between text queries and visual keys, then aggregates visual values according to these attention weights:
\begin{equation}
\mathbf{Z} = \text{Attention}(\mathbf{Q}, \mathbf{K}, \mathbf{V}_{\text{attn}}) = \text{Softmax}\left(\frac{\mathbf{Q}\mathbf{K}^T}{\sqrt{d_k}}\right)\mathbf{V}_{\text{attn}}
\end{equation}
where $d_k = 768$ is the key dimension.

The attended visual features $\mathbf{Z} \in \mathbb{R}^{N \times 768}$ are then combined with the text representations through concatenation and linear projection:
\begin{equation}
\mathbf{H}^5_{\text{fused}} = \text{LN}(\mathbf{W}_O [\mathbf{A}^5; \mathbf{Z}] + \mathbf{A}^5)
\end{equation}
where $[\cdot;\cdot]$ denotes concatenation, $\mathbf{W}_O \in \mathbb{R}^{768 \times 1536}$ projects the concatenated features back to 768 dimensions, and we apply a residual connection with layer normalization to preserve text information flow. The concatenation-then-project design allows the model to learn rich interactions between modalities, while the residual connection ensures text information is preserved when visual information is less relevant.


\subsection{Training Objective}

We train \methname{} end-to-end by maximizing the conditional log-likelihood of generating the target summary $\mathcal{S} = [s_1, s_2, ..., s_L]$ given the input text and video:
\begin{equation}
\mathcal{L} = -\sum_{i=1}^{L} \log P(s_i | s_{<i}, \mathcal{T}, \mathcal{V}; \theta)
\end{equation}
where $s_i$ is the $i$-th token of the summary, $s_{<i}$ denotes all previously generated tokens, and $\theta$ represents all model parameters including BART encoder-decoder weights, visual feature projection, cross-modal attention parameters, and temporal positional encodings. Note that CLIP visual encoder parameters remain frozen and are not included in $\theta$.

%% file: sec/4_result.tex
\section{Experimental Settings}
\label{sec:experiments}

\subsection{Dataset}

\textbf{YouCook2.} We evaluate \methname{} on YouCook2~\cite{zhou2018towards}, a large-scale instructional video dataset containing 2,000 cooking videos across 89 recipe categories. The dataset provides temporally-annotated procedural steps for each video, with an average of 7.7 steps per video describing the cooking procedure. Videos are split into 1,333 for training, 457 for validation, and 210 for testing. Videos average approximately 5 minutes in duration.

\noindent\textbf{Summary Generation.} YouCook2 provides procedural step descriptions but lacks abstractive summaries required for training summarization models. We generate summary targets using GPT-3.5-turbo, concatenating procedural step descriptions (average 69 words) and prompting for concise summaries (average 24 words), achieving summaries that are approximately 35\% of the original input length. We manually review and correct all generated summaries, addressing factual errors, excessive extractiveness, and low fluency. We retain 1,790 verified summaries for training and validation, with the 210-video test set using identically generated and verified summaries for final evaluation.

\subsection{Implementation Details}

\textbf{Data preprocessing.} For each video, we uniformly sample 50 frames and extract visual features using frozen CLIP ViT-B/32~\cite{radford2021learning}, with each frame resized to $224 \times 224$ pixels and encoded into 512-dimensional features. Input procedural steps are tokenized using BART's BPE tokenizer with a maximum sequence length of 512 tokens, truncating from the end if exceeded.

\noindent\textbf{Model Architecture.} We initialize from pre-trained BART-base~\cite{lewis2020bart} with 6 encoder layers, 6 decoder layers, and a hidden dimension of 768. For temporal modeling, we use a 2-layer Transformer encoder with 4 attention heads and a feed-forward dimension of 1024. Visual features are projected to 768 dimensions and injected into BART encoder layer 5 through cross-modal attention. Pre-trained BART parameters are fine-tuned with a learning rate $3\times10^{-5}$, while newly added layers (projection, cross-modal attention, and temporal encodings) use learning rate $1.5\times10^{-4}$.

\noindent\textbf{Training.} We train for 100 epochs using the Adam optimizer~\cite{kingma2015adam} with $\beta_1=0.9$, $\beta_2=0.999$, and weight decay $1\times10^{-5}$. We use batch size 16 with gradient accumulation over 4 steps (effective batch size 64). We apply a learning rate scheduler that decays to 95\% of the current rate every 10 epochs. We select the best checkpoint based on validation ROUGE-2 scores with early stopping patience of 10 epochs.

\noindent\textbf{Inference.} We use beam search decoding with beam size 5 and no-repeat 3-gram blocking. Decoding terminates upon generating an end-of-sequence token or reaching the 128 tokens maximum length.

\noindent\textbf{Software and Hardware.} We implement \methname{} using PyTorch 2.0~\cite{paszke2019pytorch} and conduct all experiments on a single NVIDIA RTX 4090 GPU (24GB).

\subsection{Baselines}

We compare \methname{} against text-only baselines, multimodal methods, and controlled ablations of our approach.

\noindent\textbf{Text-Only Baselines.}
\textbf{BART}~\cite{lewis2020bart}: BART-base trained only on procedural text without visual information, establishing the added value of visual features.
\textbf{T5}~\cite{raffel2020exploring}: T5-base encoder-decoder trained on text only, providing comparison across pre-trained language model architectures.

\noindent\textbf{Multimodal Baselines.}
\textbf{HA-Transformer}~\cite{libovicky2017attention,palaskar2019multimodal}: A multi-source Transformer with hierarchical attention that integrates information from different modalities into a coherent output.
\noindent\textbf{MFFG-Transformer}~\cite{liu2020multistage}: Multistage fusion with forget gate, employing a cross-fusion block and hierarchical fusion decoder for multimodal generation.
\noindent\textbf{VG-BART}~\cite{yu2021vision}: Vision-guided BART with multi-head cross-modal attention, representing a strong baseline in multimodal abstractive summarization.

\noindent\textbf{Visual Feature Comparison.}
To validate the effectiveness of CLIP features, we compare against controlled configurations:
\textbf{ResNet-152 Features}: BART with ResNet-152 features (2048-dim) from an ImageNet-pretrained model~\cite{he2016deep}, using an identical fusion mechanism, validating the effectiveness of vision-language pre-training over classification-based visual features.
\textbf{Random Visual Features}: BART with a randomly initialized visual encoder matching CLIP's architecture, verifying that improvements stem from CLIP's pre-trained representations rather than additional parameters.

All configurations use identical training procedures, hyperparameters, and fusion architectures, with only the visual feature source varying.

\subsection{Evaluation Metrics}

We evaluate generated summaries using standard abstractive summarization metrics:

\noindent\textbf{ROUGE}~\cite{lin2004rouge} measures n-gram overlap between generated and reference summaries. We report ROUGE-1 (unigram), ROUGE-2 (bigram), and ROUGE-L (longest common subsequence) F1 scores.

\noindent\textbf{BLEU}~\cite{papineni2002bleu} measures precision of n-gram matches. We report BLEU-4 scores, computed as the geometric mean of 1-4 gram precisions with a brevity penalty.

\noindent\textbf{METEOR}~\cite{banerjee2005meteor} accounts for stemming, synonymy, and paraphrasing, providing semantic-level evaluation beyond exact matches.

We compute ROUGE metrics using the \texttt{rouge-score} package (v0.1.2) and BLEU/METEOR using the \texttt{nlg-eval} toolkit. All scores are reported as percentages, with F1 scores for ROUGE and METEOR and precision scores for BLEU.

\section{Results and Analysis}
\label{sec:results}

\subsection{Main Results}

Table~\ref{tab:main_results} presents the performance comparison on YouCook2 abstractive summarization. \methname{} achieves 33.0\% ROUGE-1, 6.4\% ROUGE-2, and 25.3\% ROUGE-L, outperforming all baselines across all metrics. The consistent improvements in BLEU-4 (10.8\%) and METEOR (22.1\%) indicate benefits beyond simple n-gram overlap, extending to semantic similarity.

\begin{table}[t]
\centering
\caption{Performance comparison on YouCook2 abstractive summarization. R-1/R-2/R-L: ROUGE-1/2/L F1 scores (\%), B-4: BLEU-4 (\%), M: METEOR (\%). Best results are in \textbf{bold}, second best among all methods are \underline{underlined}.}
\label{tab:main_results}
\resizebox{\textwidth}{!}{%
\begin{tabular}{llccccc}
\toprule
\textbf{Method} & \textbf{Visual Features} & \textbf{R-1} & \textbf{R-2} & \textbf{R-L} & \textbf{B-4} & \textbf{M} \\
\midrule
\multicolumn{7}{l}{\textit{Text-Only Baselines}} \\
BART~\cite{lewis2020bart} & -- & 28.3 & 3.8 & 22.1 & 8.2 & 18.5 \\
T5~\cite{raffel2020exploring} & -- & 27.8 & 3.5 & 21.6 & 7.9 & 17.9 \\
\midrule
\multicolumn{7}{l}{\textit{Multimodal Baselines}} \\
HA-Transformer~\cite{palaskar2019multimodal} & ResNet (2048-dim) & 29.4 & 4.3 & 23.1 & 8.9 & 19.2 \\
MFFG-Transformer~\cite{liu2020multistage} & ResNet (2048-dim) & 30.1 & 4.8 & 23.6 & 9.2 & 19.8 \\
VG-BART~\cite{yu2021vision} & ResNet (2048-dim) & \underline{30.8} & \underline{5.3} & \underline{24.2} & \underline{9.7} & \underline{20.6} \\
\midrule
\multicolumn{7}{l}{\textit{Visual Feature Comparison}} \\
BART + Random Visual & Random (512-dim) & 29.1 & 4.2 & 22.8 & 8.7 & 19.1 \\
BART + ResNet-152 & ImageNet (2048-dim) & 30.5 & 5.1 & 23.9 & 9.5 & 20.3 \\
\midrule
\textbf{\methname{} (Ours)} & CLIP ViT-B/32 (512-dim) & \textbf{33.0} & \textbf{6.4} & \textbf{25.3} & \textbf{10.8} & \textbf{22.1} \\
\bottomrule
\end{tabular}%
}
\end{table}

\noindent\textbf{Comparison with Text-Only Baselines.}
Both text-only models (BART and T5) achieve comparable performance, with BART slightly outperforming T5 across all metrics. Comparing text-only BART (28.3\% ROUGE-1) with \methname{} (33.0\%), we observe a 4.7-point absolute improvement (16.6\% relative). This demonstrates that visual information provides meaningful complementary signals for summarization. 

\noindent\textbf{Comparison with Multimodal Baselines.}
Among existing multimodal methods, VG-BART achieves the strongest performance (30.8\% ROUGE-1), followed by MFFG-Transformer (30.1\%) and HA-Transformer (29.4\%). \methname{} outperforms VG-BART by 2.2 points in ROUGE-1 (7.1\% relative improvement) and 1.1 points in ROUGE-2 (20.8\% relative improvement). Notably, all these baselines use ResNet-based visual features, while \methname{} leverages CLIP's vision-language alignment. The consistent improvement across all metrics demonstrates the advantage of using semantically aligned visual representations over classification-based features for text generation tasks.

\noindent\textbf{Importance of Visual Pre-training.}
Random visual features achieve only 29.1\% ROUGE-1, merely 0.8 points above text-only BART. 
In contrast, both ResNet-152 and CLIP substantially outperform random initialization, confirming that pre-trained visual representations are essential for effective multimodal summarization.

\noindent\textbf{Vision-Language Alignment.}
\methname{} with frozen CLIP features (512-dim) achieves 33.0\% ROUGE-1 compared to ResNet-152's 30.5\%, despite using 4$\times$ lower dimensionality (512 vs. 2048 dimensions). This 2.5-point improvement demonstrates that semantic alignment with language is more valuable than raw feature capacity for text generation tasks. 
The pronounced ROUGE-2 improvement (5.1\% $\rightarrow$ 6.4\%, 25.5\% relative) suggests that vision-language features particularly help generate appropriate action-object combinations common in cooking instructions, such as ``coat chicken with sauce'' or ``fold in egg whites.''

%% file: sec/5_ablation.tex
\subsection{Ablation Studies}

To analyze the impact of key design choices in \methname{}, we conduct ablation studies on three critical components: fusion layer location, CLIP encoder training strategy, and temporal sampling density. Table~\ref{tab:ablations} summarizes the results.

\begin{table}[t]
\centering
\caption{Ablation studies analyzing key design choices in \methname{}. All experiments use CLIP ViT-B/32 visual features with identical training procedures. Best results are in \textbf{bold}.}
\label{tab:ablations}
\resizebox{\textwidth}{!}{%
\begin{tabular}{lcccclcclccc}
\toprule
 & \multicolumn{4}{c}{\textit{Fusion Layer}} & & \multicolumn{2}{c}{\textit{CLIP Strategy}} & & \multicolumn{3}{c}{\textit{Sampled Frames}} \\
\cmidrule{2-5} \cmidrule{7-8} \cmidrule{10-12}
\textbf{Metric} & L3 & L4 & L5 & L6 & & Fine-tuned & Frozen & & 25 & 50 & 100 \\
\midrule
R-1 & 31.8 & 32.4 & \textbf{33.0} & 32.1 & & 32.3 & \textbf{33.0} & & 32.1 & \textbf{33.0} & 33.1 \\
R-2 & 5.7  & 6.0  & \textbf{6.4}  & 5.9  & & 6.0  & \textbf{6.4}  & & 5.8  & \textbf{6.4}  & 6.5  \\
R-L & 24.2 & 24.7 & \textbf{25.3} & 24.5 & & 24.8 & \textbf{25.3} & & 24.4 & \textbf{25.3} & 25.4 \\
\bottomrule
\end{tabular}%
}
\end{table}

\noindent\textbf{Fusion Layer Location.} 
Mid-layer fusion at encoder layer 5 achieves 33.0\% ROUGE-1, compared to 31.8\% for early fusion (layer 3) and 32.1\% for late fusion (layer 6). This validates our design principle: processing textual information through early encoder layers (1--4) before incorporating visual guidance at the mid-layer allows the final layer (6) to refine multimodal representations before decoding. 

\noindent\textbf{CLIP Encoder Strategy.} 
Freezing CLIP (33.0\% ROUGE-1) outperforms fine-tuning (32.3\%), demonstrating that preserving CLIP's pre-trained vision-language alignment is more valuable than task-specific adaptation. Fine-tuning on our limited training set (1,333 videos) appears to cause catastrophic forgetting of the general visual-semantic understanding learned from 400M image-text pairs. 

\noindent\textbf{Number of Sampled Frames.} 
Using 50 frames achieves 33.0\% ROUGE-1, substantially outperforming 25 frames (32.1\%) but performing nearly identically to 100 frames (33.1\%, only +0.1 point). This indicates that 50 uniformly sampled frames adequately capture the temporal progression of cooking videos averaging 5 minutes in duration, with diminishing returns from denser sampling. Since 100 frames double computational cost for minimal gain, we select 50 frames.

\subsection{Qualitative Analysis}

Table~\ref{tab:examples} presents example summaries from different configurations on the YouCook2 test set, illustrating qualitative differences in summarization quality.

\begin{table}[t]
\centering
\small
\caption{Example summaries generated by different configurations on a YouCook2 test video. Key procedural elements are \underline{underlined} in the reference for comparison.}
\label{tab:examples}
\begin{tabular}{p{0.99\linewidth}}
\toprule
\textbf{Input (Procedural Steps):} \\
\textit{Mix sesame seeds with chicken breast. Add cooking oil, soy sauce, and salt and mix everything. Add 2 tbsp cornstarch and let it marinate. Combine chicken stock, dark soy sauce, cooking wine, hoisin sauce, black vinegar, sugar, and sesame oil. Mix corn starch and water together. Coat the chicken pieces with flour and fry them in oil. Fry minced garlic and roasted red chili and add the sauce. Add the corn starch and water mixture. Add the fried chicken and stir to coat well.} \\
\midrule
\textbf{Reference Summary:} \\
\textit{\underline{Marinate} sesame chicken in soy sauce, salt, and cornstarch, then \underline{fry} and \underline{coat} with a savory sauce of chicken stock, hoisin, and spices.} \\
\midrule
\textbf{Text-Only BART:} \\
\textit{Mix sesame seeds with chicken breast and fry in oil until coated well with sauce.} \\
\midrule
\textbf{BART + ResNet-152:} \\
\textit{Marinate chicken in soy sauce and cornstarch, then fry with garlic and red chili in oil.} \\
\midrule
\textbf{\methname{} (Ours):} \\
\textit{Marinate chicken with sesame seeds and soy sauce, fry until golden, and coat with a hoisin sauce mixture.} \\
\bottomrule
\end{tabular}
\end{table}

\noindent\textbf{Text-only BART} oversimplifies the procedure, missing key ingredients (cornstarch, hoisin sauce) and the marinating step entirely. 
\noindent\textbf{BART + ResNet-152} incorporates more details by leveraging visual information, correctly identifying the marinating step and specific ingredients (garlic, red chili), but fails to capture the final coating step with hoisin sauce.

\noindent\textbf{\methname{}} captures all three major procedural phases (marinate $\rightarrow$ fry $\rightarrow$ coat) and key ingredients (sesame seeds, soy sauce, hoisin sauce), producing a summary that closely matches the reference in both content coverage and abstraction level. Notably, \methname{} generates ``fry until golden,'' a visual description not present in the input text but observable in video frames. This demonstrates how CLIP's vision-language alignment enables identifying salient visual states: the model understands ``golden'' as a semantically meaningful cooking indicator aligned with the concept of properly fried food.

%% file: sec/6_conclusion.tex
\section{Conclusion}
We presented \methname{}, a multimodal framework for instructional video summarization that integrates frozen CLIP visual features with BART through explicit temporal modeling and dimension-adaptive cross-modal fusion. On YouCook2, \methname{} achieves 33.0\% ROUGE-1, outperforming ResNet-based approaches (30.5\%) despite using 4$\times$ lower-dimensional representations, and our ablations show that freezing CLIP outperforms fine-tuning while mid-layer fusion provides the most effective cross-modal integration.

These results yield three design principles for multimodal generation: prioritize vision-language alignment over feature dimensionality, freeze pre-trained encoders rather than adapt them on limited data, and inject visual information at intermediate layers for balanced modality integration. Freezing the visual encoder also reduces computational cost, making multimodal summarization more accessible in resource-constrained settings. More broadly, our findings suggest that the value of large-scale vision-language pre-training lies in faithful preservation of its learned alignment rather than task-specific adaptation, with implications beyond summarization. Future work will explore stronger vision-language encoders, multi-granularity temporal modeling, and extension to broader instructional domains beyond cooking.